\pdfoutput=1
\documentclass[11pt]{article}
\usepackage{naacl2021}
\usepackage{times}
\usepackage{latexsym}
\usepackage{comment}
\usepackage{url}
\usepackage{amsmath}
\usepackage{CJKutf8}
\usepackage[boxed]{algorithm}
\usepackage{caption}
\usepackage{algpseudocode}
{\begin{itemize}[topsep=0pt, partopsep=0pt] 
	\setlength{\itemsep}{0pt}%
	\setlength{\parskip}{0pt}%
}%
{\end{itemize}}
\usepackage{booktabs}
\usepackage{subcaption}
\usepackage{xspace}
\usepackage{amstext}
\usepackage{booktabs}
\usepackage{graphicx}
\usepackage{tabulary}
\usepackage{graphicx}
\usepackage{enumitem}
\usepackage{color}
\usepackage{lipsum}
\usepackage{amsfonts}

\usepackage{relsize}
\usepackage{bbm}
\usepackage{times}
\usepackage{epsfig}
\usepackage{amssymb}
\usepackage{makecell}
\usepackage{multirow}
\usepackage{bm}
\usepackage{placeins}

\title{Cross-Lingual Named Entity Recognition Using Parallel Corpus: A New Approach Using XLM-RoBERTa Alignment}

\author{Bing Li \\
	Microsoft \\
	\texttt{libi@microsoft.com} \\\And
	Yujie He \\
	Microsoft \\
	\texttt{yujh@microsoft.com} \\\And
    Wenjin Xu \\
	Microsoft \\
	\texttt{wenjinxu@microsoft.com} \\}

\begin{document}
\maketitle

\begin{abstract}
We propose a novel approach for cross-lingual Named Entity Recognition (NER) zero-shot transfer using parallel corpora. We built an entity alignment model on top of XLM-RoBERTa to project the \emph{entities} detected on the English part of the parallel data to the target language sentences, whose accuracy surpasses all previous unsupervised models. With the alignment model we can get pseudo-labeled NER data set in the target language to train task-specific model. Unlike using translation methods, this approach benefits from natural fluency and nuances in target-language original corpus. We also propose a modified loss function similar to focal loss but assigns weights in the opposite direction to further improve the model training on noisy pseudo-labeled data set. We evaluated this proposed approach over 4 target languages on benchmark data sets and got competitive F1 scores compared to most recent SOTA models. We also gave extra discussions about the impact of parallel corpus size and domain on the final transfer performance.
\end{abstract}

\section{Introduction}
Named entity recognition (NER) is a fundamental task in natural language processing, which seeks to classify words in a sentence to predefined semantic types. Due to its nature that the ground truth label exists at word level, supervised training of NER models often requires large amount of human annotation efforts. In real-world use cases where one needs to build multi-lingual models, the required human labor scales at least linearly with number of languages, or even worse for low resource languages. Cross-Lingual transfer on Natural Language Processing(NLP) tasks has been widely studied in recent years~\cite{conneau-etal-2018-xnli, kim-etal-2017-cross, Ni2017, xie-etal-2018-neural, Ni2019, WuDredze2019, bari2019zero, jain2019entity}, in particular zero-shot transfer which leverages the advances in high resource language such as English to benefit other low resource languages. In this paper, we focus on the cross-lingual transfer of NER task, and more specifically using parallel corpus and pretrained multilingual language models such as mBERT~\cite{multilingualBERTmd} and XLM-RoBERTa (XLM-R)~\cite{lample2019cross, Conneau2020}.

Our motivations are threefold. (1) Parallel corpus is a great resource for transfer learning and is rich between many language pairs. Some recent research focus on using completely unsupervised machine translations (e.g. word alignment~\cite{conneau2017word}) for cross-lingual NER, however inaccurate translations could harm the transfer performance. For example in the word-to-word translation approach, word ordering may not be well represented during translations, such gaps in translation quality may harm model performance in down stream tasks. (2) A method could still provide business value-add even if it only works for some major languages that have sufficient parallel corpus as long as it has satisfying performance. It is a common issue in industry practices where there is a heavily customized task that need to be extended into major markets but you do not want to annotate large amounts of data in other languages. (3) Previous attempts using parallel corpus are mostly heuristics and statistical-model based~\cite{jain2019entity, xie-etal-2018-neural, Ni2017}. Recent breakthroughs in multilingual language models have not been applied to such scenarios yet. Our work bridges the gap and revisits this topic with new technologies.

We propose a novel semi-supervised method for the cross-lingual NER transfer, bridged by parallel corpus. First we train an NER model on source-language data set - in this case English - assuming that we have labeled task-specific data. Second we label the English part of the parallel corpus with this model. Then, we project those recognized entities onto the target language, i.e. label the span of the same entity in target-language portion of the parallel corpus. In this step we will leverage the most recent XLM-R model~\cite{lample2019cross, Conneau2020}, which makes a major distinction between our work and previous attempts. Lastly, we use this pseudo-labeled data to train the task-specific model in target language directly. For the last step we explored the option of continue training from a multilingual model fine-tuned on English NER data to maximize the benefits of model transfer. We also tried a series of methods to mitigate the noisy label issue in this semi-supervised approach.

The main contributions of this paper are as follows:

\begin{itemize}
	\item We leverage the powerful multilingual model XLM-R for entity alignment. It was trained in a supervised manner with easy-to-collect data, which is in sharp contrast to previous attempts that mainly rely on unsupervised methods and human engineered features.
	\item Pseudo-labeled data set typically contains lots of noise, we propose a novel loss function inspired by the focal loss~\cite{focal}. Instead of using native focal loss, we went the opposite direction by weighting hard examples less as those are more likely to be noise.
	\item By leveraging existing natural parallel corpus we got competitive F1 scores of NER transfer on multiple languages. We also tested that the domain of parallel corpus is critical in an effective transfer.
\end{itemize}

\section{Related Works}
\label{section:work}
There are different ways to conduct zero-shot multilingual transfer. In general, there are two categories, model-based transfer and data-based transfer. Model-based transfer often use source language to train an NER model with language independent features, then directly apply the model to target language for inference~\cite{wu2019beto, wu2020enhanced}. Data-based transfer focus on combining source language task specific model, translations, and entity projection to create weakly-supervised training data in target language. Some previous attempts includes using annotation projection on aligned parallel corpora, translations between a source and a target langauge~\cite{Ni2017, ehrmann11}, or to utilize Wikipedia hyperlink structure to obtain anchor text and context as weak labels~\cite{alrfou15, TsaiMaRo16}. Different variants exist in annotation projection, e.g. Ni et al.~\cite{Ni2017} used maximum entropy alignment model and data selection to project English annotated labels to parallel target language sentences. Some other work used bilingual mapping combined with lexical heuristics or used embedding approach to perform word-level translation with which naturally comes the annotation projection~\cite{D17-1268, xie-etal-2018-neural, jain2019entity}. This kind of translation + projection approach is used not just in NER, but in other NLP tasks as well such as relation extraction~\cite{Faruqui:2015naaclshort}. There are obvious limitations to the translation + projection approach, word or phrase-based translation makes annotation projection easier, but sacrifices the native fluency and language nuances. In addition, orthographic and phonetic based features for entity matching may only be applicable to languages that are alike, and requires extensive human engineered features. To address these limitations, we proposed a novel approach which utilizes machine translation training data and combined with pretrained multilingual language model for entity alignment and projection.

\section{Model Design}
\label{section:model}
We will demonstrate the entity alignment model component and the full training pipeline of our work in following sections.

\subsection{Entity Alignment Model}

Translation from a source language to target language may break the word ordering therefore an alignment model is needed to project entities from source language sentence to target language, so that the labels from source language can be zero-shot transferred. In this work, we use XLM-R~\cite{lample2019cross, Conneau2020} series of models which introduced the translation language model(TLM) pretraining task for the first time. TLM trains the model to predict a masked word using information from both the context and the parallel sentence in another language, making the model acquire great cross-lingual and potentially alignment capability.

\begin{figure}[!h]
	\centering
	\includegraphics[width=1\linewidth]{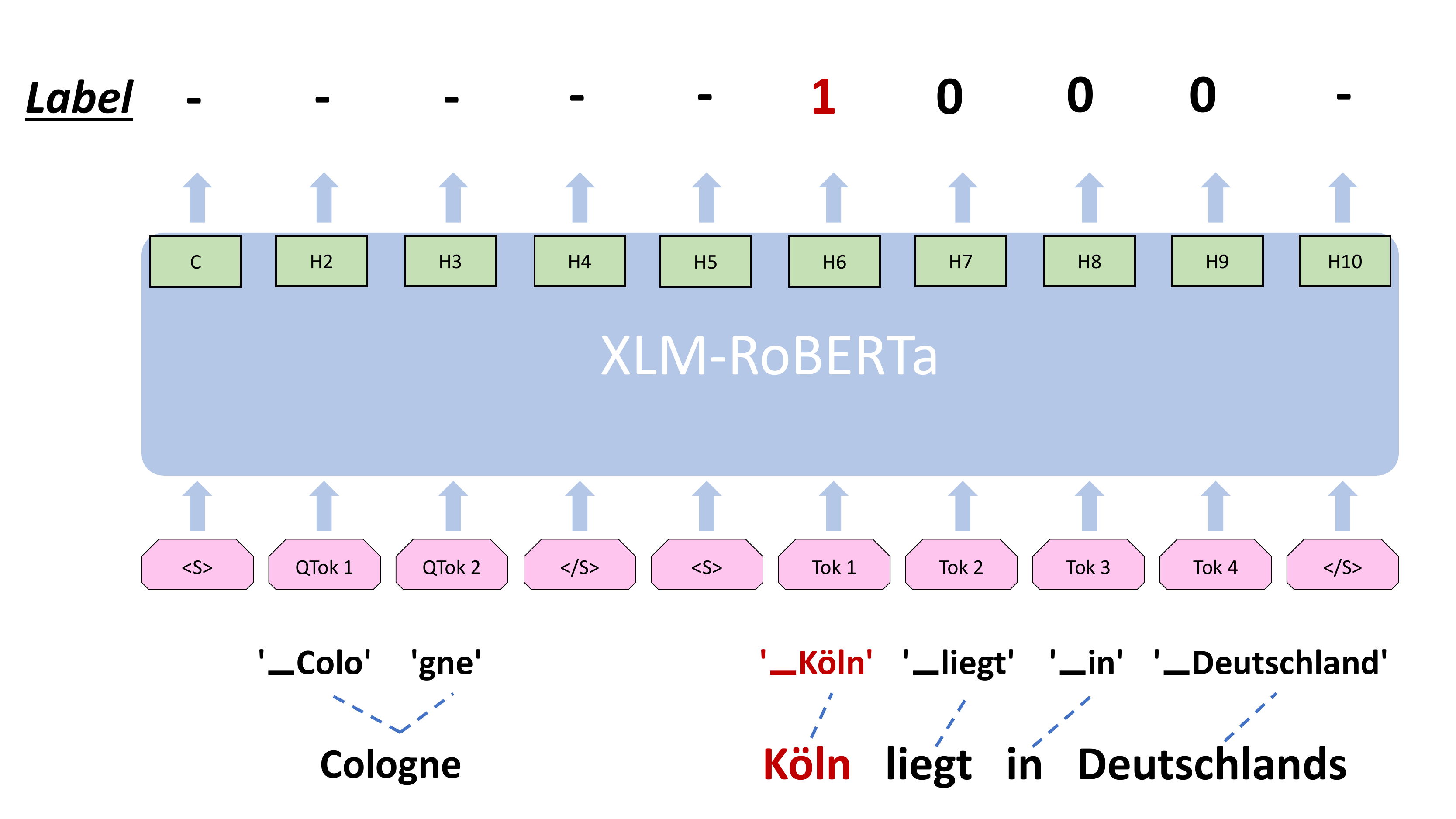}
	\caption{\textbf{Entity alignment model}. The query entity on the left is 'Cologne', and the German sentence on the right is 'Köln liegt in Deutschlands', which is 'Cologne is located in Germany' in English, and 'Köln' is the German translation of 'Cologne'. The model predicts the word span aligned with the query entity.}
	\label{fig:alignmentNetwork}
\end{figure}

Our alignment model is constructed by concatenating the English name of the entity and the target language sentence, as segment A and segment B of the input sequence respectively. For token level outputs of segment B, we predict 1 if it is inside the translated entity, 0 otherwise. This formulation transforms the entity alignment problem into a token classification task. An implicit assumption is that the entity name will still be a consecutive phrase after translation. The model structure is illustrated in Fig~\ref{fig:alignmentNetwork}.

\subsection{Cross-Lingual Transfer Pipeline}

\begin{figure*}[t]
	\centering
	\includegraphics[width=14.5cm]{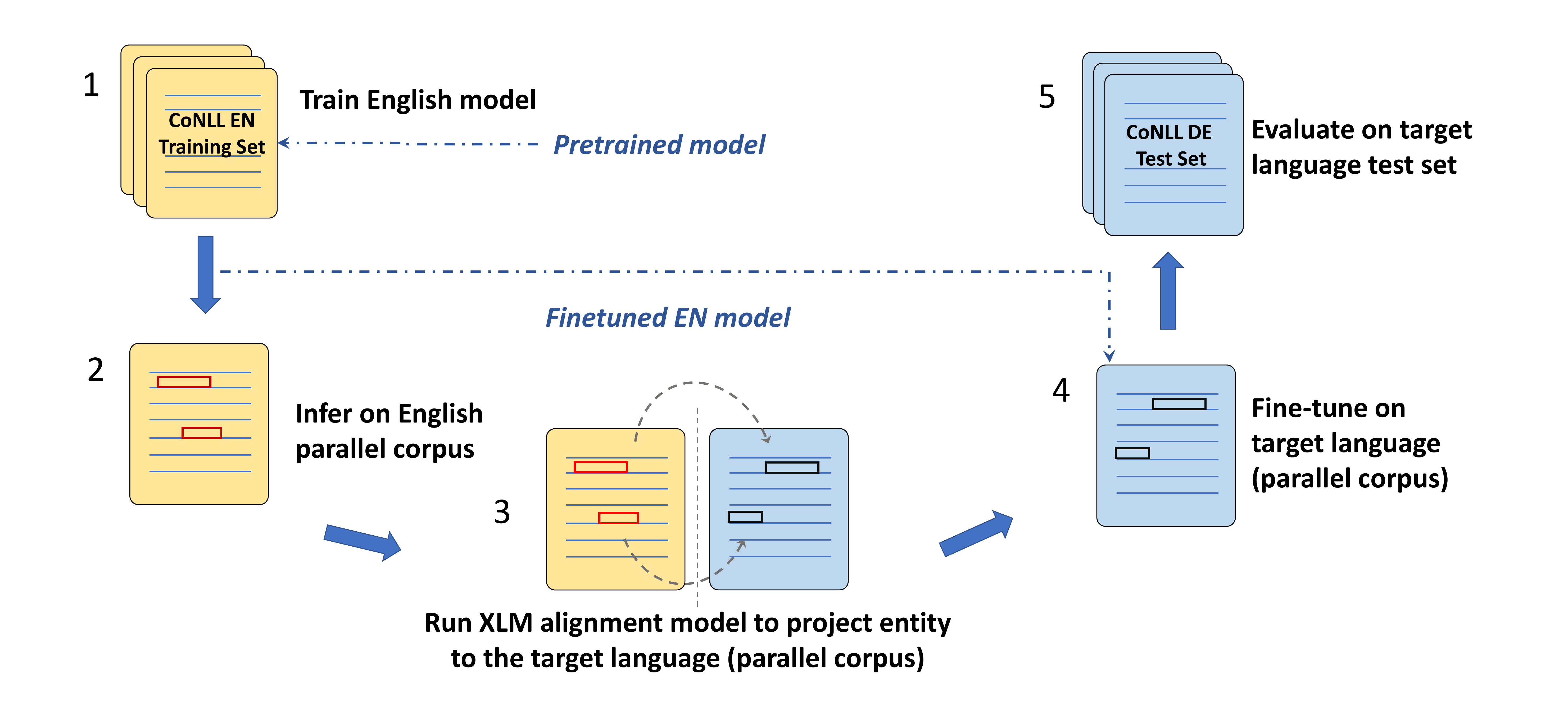}
	\caption{\textbf{Training Pipeline Diagram}. Yellow pages represent English documents while light blue pages represent German documents. Step 1 and 5 used original CoNLL data for train and test respectively; step 2, 3, and 4 used machine translation data from OPUS website. Pretrained model is either mBert or XLM-R. The final model was first fine-tuned on English NER data set then fine-tuned on target language pseudo-labeled NER data set.}
	\label{fig:trainingpipeline}
\end{figure*}

Fig~\ref{fig:trainingpipeline} shows the whole training/evaluation pipeline which includes 5 stages: (1) Fine-tune pretrained language model on CoNLL2003~\cite{conll2003ner} to obtain English NER model; (2) Infer labels for English sentences in the parallel corpus; (3) Run the entity alignment model from the previous subsection and find corresponding detected English entities in the target language, failed examples are filtered out during the alignment process; (4) Fine-tune the multilingual model with data generated from step (3); (5) Evaluate the new model on the target language test sets.

\section{Experiments And Discussions}
\label{section:experiment}

\subsection{Parallel Corpus}
In our method, we leveraged the availability of large-scale parallel corpus to transfer the NER knowledge obtained in English to other languages. Existing parallel corpora is easier to obtain than annotated NER data. We used parallel corpus crawled from the OPUS website~\cite{TIEDEMANN12}. In our experiments, we used the following data sets:

\begin{itemize}
	\item \textbf{Ted2013}: consists of volunteer transcriptions and translations from the TED web site and was created as training data resource for the International Workshop on Spoken Language Translation 2013.
	\item \textbf{OpenSubtitles}: a new collection of translated movie subtitles that contains 62 languages.~\cite{OpenSubtitles2016}
	\item \textbf{WikiMatrix}: Mined parallel sentences from Wikipedia in different languages. Only pairs with scores above 1.05 are used.~\cite{WikiMatrix}
	\item \textbf{UNPC}: Manually translated United Nations documents from 1994 to 2014.~\cite{UNPC}
	\item \textbf{Europarl}: A parallel corpus extracted from the European Parliament web site.~\cite{Europarl}
	\item \textbf{WMT-News}: A parallel corpus of News Test Sets provided by WMT for training SMT that contains 18 languages.\footnote{http://www.statmt.org/wmt19/translation-task.html}
	\item \textbf{NewsCommentary}: A parallel corpus of News Commentaries provided by WMT for training SMT, which contains 12 languages.\footnote{http://opus.nlpl.eu/News-Commentary.php}
	\item \textbf{JW300}: Mined, parallel sentences from the magazines Awake! and Watchtower.~\cite{jw300}
\end{itemize}

In this work, we focus on 4 languages, German, Spanish, Dutch and Chinese. We randomly select data points from all data sets above with equal weights. There might be slight difference in data distribution between languages due to data availability and relevance.

\subsection{Alignment Model Training}
The objective of alignment model is to find the entity from a foreign paragraph given its English name. We feed the English name and the paragraph as segment A and B into the XLM-R model~\cite{lample2019cross, Conneau2020}. Unlike NER task, the alignment task has no requirement for label completeness, since we only need one entity to be labeled in one training example. The training data set can be created from Wikipedia documents where anchor text in hyperlinks naturally indicate the location of entities and one can get the English entity name via linking by Wikipedia entity Id. An alternative to get the English name for mentions in another language is through state-of-the-art translation system. We took the latter approach to make it simple and leveraged Microsoft's Azure Cognitive Service to do the translation.

During training, we also added negative examples with faked English entities which did not appear in the other language's sentence. The intuition is to force model to focus on English entity (segment A) and its translation instead of doing pure NER and picking out any entity in other language's sentence (segment B). We also added examples of noun phrases or nominal entities to make the model more robust.

We generated a training set of 30K samples with 25$\%$ negatives for each language and trained a XLM-R-large model~\cite{Conneau2020} for 3 epochs with batch size 64. The initial learning rate is 5e-5 and other hyperparameters were defaults from HuggingFace Transformers library for the token classification task. The precision/recall/F1 on the reserved test set reached 98$\%$. The model training was done on 2 Tesla V100 and took about 20 minutes.

\subsection{Cross-Lingual Transfer}

\begin{table*}[t]
	\small
	\begin{center}
		\begin{tabular}{lrrrrrrrrrrrrrrr}
			\toprule
			&&\textbf{DE}&&&&\textbf{ES}&&&&\textbf{NL}&&&&\textbf{ZH}&\\
			\hline
			\textbf{Model} & \textbf{P} & \textbf{R} & \textbf{F1} && \textbf{P} & \textbf{R} & \textbf{ F1} && \textbf{P} & \textbf{R} & \textbf{ F1} && \textbf{P} & \textbf{R} & \textbf{F1}\\
			\hline
			\citet{bari2019zero}& - & - & 65.24 && - & - & 75.9 && - & - & 74.6 && - & - & - \\
			\citet{wu2019beto}& - & - & 71.1 && - & - & 74.5 && - & - & 79.5 && - & - & - \\
			\citet{moon2019lingua}& - & - & 71.42 && - & - & 75.67 && - & - & 80.38 && - & - & - \\
			\citet{wu2020enhanced}& - & - & 73.16 && - & - & 76.75 && - & - & 80.44 && - & - & -\\
			\citet{wu2020single}& - & - & 73.22 && - & - & 76.94 && - & - & 80.89 && - & - & -\\
			\citet{wu2020unitrans}& - & - & \textbf{74.82} && - & - & \textbf{79.31} && - & - & \textbf{82.90} && - & - & -\\
			\hline
			\textbf{Our Models}&  &  &  &&  & &  && &  &  && & &\\
			\hline
			mBERT zero-transfer & 67.6 & 77.4 & 72.1 && 72.4 &  78.2 & 75.2 && 77.8 & 79.3 & 78.6 && 64.1 & 65.0 & 64.6 \\
			mBERT fine-tune & 73.1 & 76.2 & \textbf{74.6} && 77.7 & 77.6 & \textbf{77.6} && 80.5 & 76.7 & 78.6 && 80.8 & 63.3 & \textbf{71.0} \\
			&  & & \textbf{\scriptsize{(+2.5)}} &&  & & \textbf{\scriptsize{(+2.4)}} &&  & & \textbf{\scriptsize{(+0.0)}} && & & \textbf{\scriptsize{(+6.4)}} \\
			\hline
			XLM-R zero-transfer & 67.9 & 79.8 & 73.4 && 79.8 & 81.9 & \textbf{80.8} && 82.2 & 80.3 & \textbf{81.2} && 68.7 & 65.5 & 67.1 \\
			XLM-R fine-tune & 75.5 & 78.4 & \textbf{76.9} && 76.9 & 81.0 & 78.9 && 81.2 & 78.4 & 79.7 && 77.8 & 65.9 & \textbf{71.3} \\
			&  & & \textbf{\scriptsize{(+3.5)}} &&  & & \textbf{\scriptsize{(-1.9)}} &&  & & \textbf{\scriptsize{(-1.5)}} && & & \textbf{\scriptsize{(+4.2)}} \\
			\toprule
		\end{tabular}
	\end{center}
	\caption{\textbf{Cross-Lingual Transfer Results on German, Spanish, Dutch and Chinese}: Experiments are done with both mBERT and XLM-RoBERTa model. For each of them we compared zero-transfer result (trained on CoNLL2003 English only) and fine-tune result using zero-transfer pseudo-labeled target language NER data. Test sets for German, Spanish, Dutch are from CoNLL2003 and CoNLL2002, and People's Daily data set for Chinese.}
	\label{tab:transferresults}
\end{table*}

We used the CoNLL2003~\cite{conll2003ner} and CoNLL2002\footnote{http://lcg-www.uia.ac.be/conll2002/ner/} data sets to test our cross-lingual transfer method for German, Spanish and Dutch. We ignored the training sets in those languages and only evaluated our model on test sets. For Chinese, we used People's Daily\footnote{https://github.com/zjy-ucas/ChineseNER} as the major evaluation set, and we also reported numbers on MSRA~\cite{levow-2006-third} and Weibo~\cite{peng-dredze-2015-named} data sets in the next section. One notable difference for People's Daily data set is that it only covers three entity types, LOC, ORG and PER, so we suppressed the MISC type from English during the transfer by training English NER model with 'MISC' marked as 'O'.

To enable cross-lingual transfer, we first trained an English teacher model using the CoNLL2003 EN training set with XLM-R-large as the base model. We trained with focal loss~\cite{focal} for 5 epochs. We then ran inference with this model on the English part of the parallel data. Finally, with the alignment model, we projected entity labels onto other languages. To ensure the quality of target language training data, we discarded examples if any English entity failed to map to tokens in the target language. We also discarded examples where there are overlapping target entities because it will cause conflicts in token labels. Furthermore, when one entity is mapped to multiple, we only keep the example if all the target mention phrases are the same. This is to accommodate the situation where same entity is mentioned more than once in one sentence.

As the last step, we fine-tuned the multilingual model pre-trained on English data set with lower n(0, 3, 6, etc.) layers frozen, on the target language pseudo-labeled data. We used both mBERT~\cite{devlin-etal-2019-bert, multilingualBERTmd} and XLM-R\cite{Conneau2020} with about 40K training samples for 1 epoch. The results are shown in Table~\ref{tab:transferresults}. All the inference, entity projection and model training experiments are done on 2 Tesla V100 32G gpus and the whole pipeline takes about 1-2 hours. All numbers are reported as an average of 5 random runs with the same settings.

For loss function we used something similar to focal loss~\cite{focal} but with opposite weight assignment. The focal loss was designed to weight hard examples more. This intuition holds true only when training data is clean. In some scenarios such as the cross-lingual transfer task, the pseudo-labeled training data contains lots of noise propagated from upper-stream of the pipeline, in which case, those 'hard' examples are more likely to be just errors or outliers and could hurt the training process. We went the opposite direction and lowered their weights instead, so that the model could focus on less noisy labels. More specifically, we added weight $(1+p_t)^{\gamma}$ instead of $(1-p_t)^{\gamma}$ on top of the regular cross entropy loss, and for the hyper-parameter $\gamma$ we experimented with values from 1 to 5 and the value of 4 it works best.

From Table~\ref{tab:transferresults}, we can see for mBERT model, fine-tuning with pseudo-labeled data has significant effects on all languages except NL. The largest improvement is in Chinese, 6.4$\%$ increase in F1 on top of the zero-transfer result, this number is 2.5$\%$ for German and 2.4$\%$ for Spanish. The same experiment with XLM-R model shows a different pattern, F1 increased 3.5$\%$ for German but dropped a little bit on Spanish and Dutch after fine-tuning. For Chinese, we see a comparable improvement with mBERT of 4.2$\%$. The negative result on Spanish and Dutch is probably because XLM-R has already had very good pretraining and language alignment for those European languages, which can be seen from the high zero-transfer numbers, therefore a relatively noisy data set did not bring much gain. On the contrary, Chinese is a relatively distant language from the perspective of linguistics so that the add-on value of task specific fine-tuning with natural data is larger.

Another pattern we observed from Table~\ref{tab:transferresults} is that across all languages, the fine-tuning step with pseudo-labeled data is more beneficial to precision compared with recall. We observed a consistent improvement in precision but a small drop in recall in most cases.

\section{Discussions}
\label{section:further}

\subsection{Impact of the amount of parallel data}
One advantage of the parallel corpora method is high data availability compared to the supervised approach. Therefore a natural question next is whether more data is beneficial for the cross-lingual transfer task. To answer this question, we did a series of experiments with varying number of training examples ranging from 5K to 200K, and the model F1 score increases with the amount of data at the beginning and plateaued around 40K. All the numbers displayed in Table~\ref{tab:transferresults} are reported for training on a generated data set of size around 40K (sentences). One possible explanation for the plateaued performance might be due to the propagated error in the pseudo-labeled data set. Domain mismatch may also limit the effectiveness of transfer between languages. More discussions on this topic in the next section. 

\subsection{Impact of the domain of parallel data}

\begin{figure}[!h]
	\centering
	\includegraphics[width=1\linewidth]{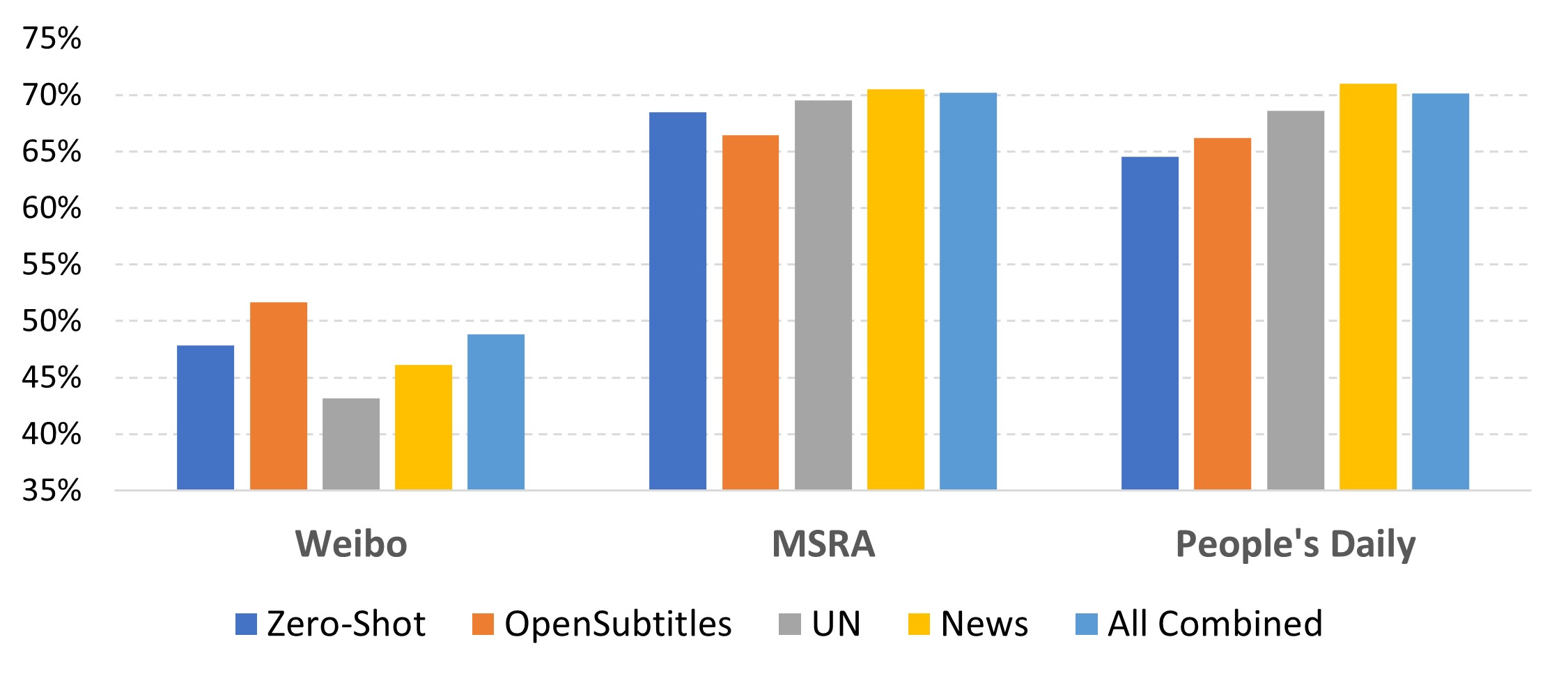}
	\caption{F1 scores evaluated on three data sets using different domains' parallel data. Blue column on the left is the result of zero-shot model transfer. To its right are F1 scores for 3 different domains and all domains combined.}
	\label{fig:domain}
\end{figure}

Learnings from machine translation community showed that the quality of neural machine translation models usually strongly depends on the domain they are trained on, and the performance of a model could drop significantly when evaluated on a different domain~\cite{koehn-knowles-2017-six}. Similar observation can be used to explain challenges of the NER cross-lingual transfer. In NER transfer, the first domain mismatch comes from the natural gap in entity distributions between different language corpus. Many entities only live inside the ecosystem of a specific group of languages and may not be translated naturally to others. The second domain mismatch is between the parallel corpora and the NER data set. The English model might not have good domain adaptation ability and could perform well on the CoNLL2003 data set but poorly on the parallel data set.

\begin{table}[h]
	\scriptsize
	\begin{center}
		\begin{tabular}{c|c|c|c|c}
			\toprule
			\textbf{Domain} & \textbf{PER} & \textbf{ORG} & \textbf{LOC} & \textbf{All}\\
			\hline
			OpenSubtitles & 24,036 & 3,809 & 5,196 & 33,041   \\
			\hline
			UN	& 1,094 & 25,875 & 12,718 & 39,687  \\
			\hline
			News & 10,977 & 9,568 & 28,168 & 48,713 \\
			\hline
			\textbf{All Domains Combined}	& 10,454 & 14,269 & 17,412 & 42,135 \\
			\toprule
		\end{tabular}
	\end{center}
	\caption{Entity Count by type in the pseudo-labeled Chinese NER training data set. We listed multiple domains that were extracted from different parallel data source. And AllDomains is a combination of all resources.}
	\label{tab:typedistribution}
\end{table}

To study the impact of the domain of parallel data on transfer performance, we did an experiment on Chinese using parallel data from different domains. We picked three representative data sets from OPUS~\cite{TIEDEMANN12}, OpenSubtitles, UN(contains UN and UNPC), News(contains WMT and News-Commentary) and another one with all these three combined. OpenSubtitles is from movie subtitles and language style is informal and oral. UN is from united nation reports and language style is more formal and political flavored. News data is from newspaper and content is more diverse and closer to CoNLL data sets. We evaluated the F1 on three Chinese testsets, Weibo, MSRA and People's Daily, where Weibo contains messages from social media, MSRA is more formal and political, and People's Daily data set are newspaper articles. 

From Fig~\ref{fig:domain} we see OpenSubtitles performs best on Weibo but poorly on the other two. On the contrary UN performs worst on Weibo but better on the other two. News domain performs the best on People's Daily, which is also consistent with one's intuition because they are both from newspaper articles. All domains combined approach has a decent performance on all three testsets.

Different domains of data have quite large gap in the density and distribution of entities, for example OpenSubtitles contains more sentences that does not have any entity. In the experiments above we did filtering to keep the same ratio of 'empty' sentences among all domains. We also examined the difference in type distributions. In Tab~\ref{tab:typedistribution}, we calculated entity counts by type and domain. OpenSubtitles has very few ORG and LOC entities, whereas UN data has very few PER data. News and All domain data are more balanced.

\begin{figure}[!h]
	\centering
	\includegraphics[width=1\linewidth]{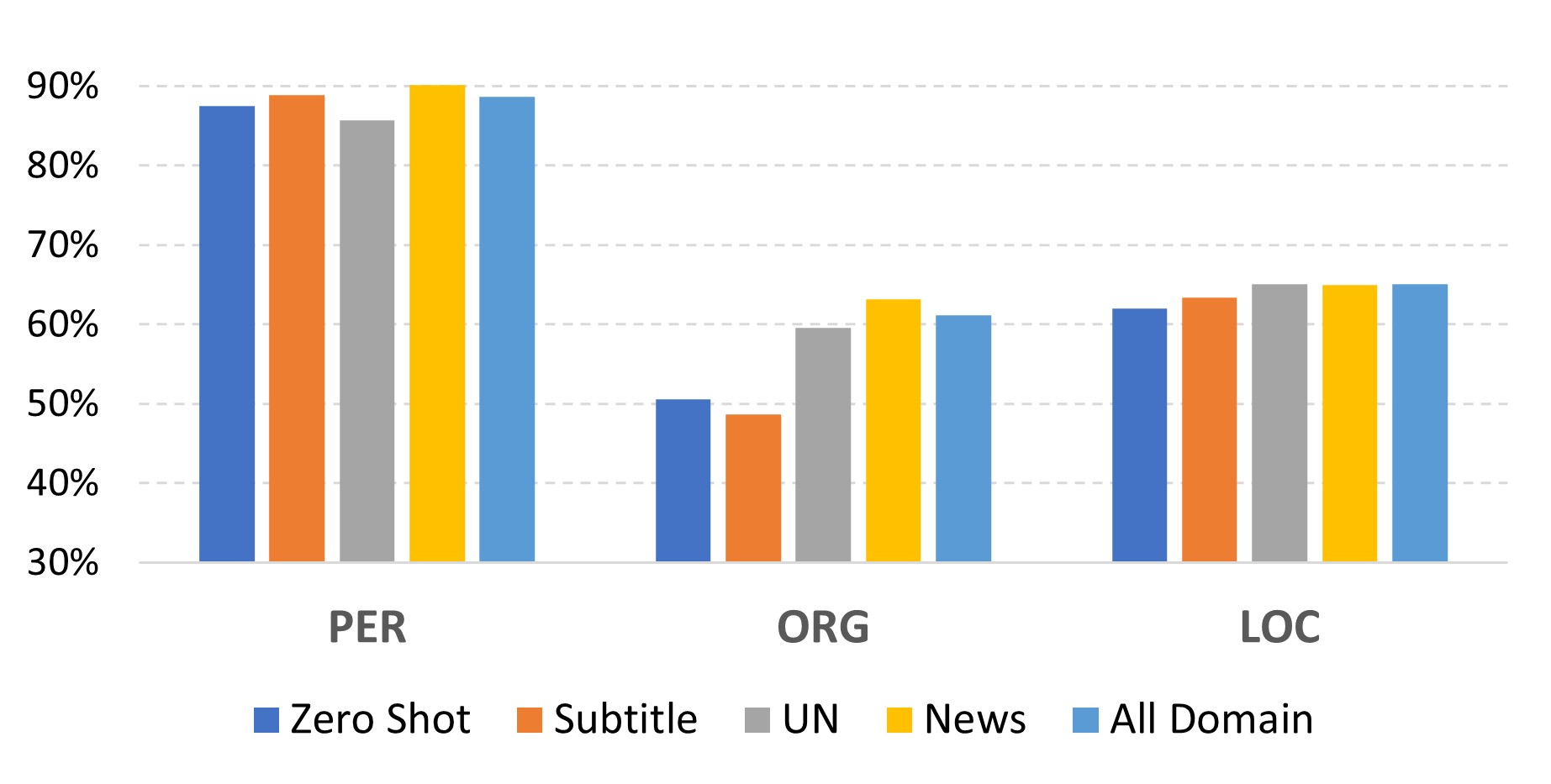}
	\caption{F1 score by type evaluated on People's Daily data set. The same as Fig~\ref{fig:domain}, we compare the results using different domains of parallel data for the NER transfer.}
	\label{fig:bytype}
\end{figure}

In Fig~\ref{fig:bytype}, we show evaluation on People's Daily by type. We want to understand how does parallel data domain impacts transfer performance for different entity types. News data has the best performance on all types, and Subtitles has very bad result on ORG. All these observations are consistent with the type distribution in Table~\ref{tab:typedistribution}.

\section{Ablation Study}
\label{section:ablation}
To better understand the contribution of each stage in the training process, we conducted an ablation study on German data set with mBERT model. We compared 6 different settings: (1) approach proposed in this work, i.e. fine-tune the English model with pseudo-labeled data with the new loss, denoted as re-weighted (RW) loss; (2) zero-transfer is direct model transfer which was trained only on English NER data; (3) fine-tune the English model with pseudo-labeled data with regular cross-entropy (CE) loss; (4) Skip fine-tune on English and directly fine-tune mBERT on pseudo-labeled data. (5)(6) fine-tune the model directly with mixed English and pseudo-labeled data simultaneously with RW and CE losses respectively.

\begin{table}[t]
	\small
	\begin{center}
		\begin{tabular}{lrrr}
			\toprule
			& \textbf{Test P} & \textbf{Test R} & \textbf{F1}\\
			\hline
			\textbf{Sequential fine-tune with RW} & \textbf{73.1} &  76.2 & \textbf{74.6}    \\
			Zero-transfer & 67.6 & \textbf{77.4} & 72.1   \\
			Sequential fine-tune with CE  & 71.2 & 74.3 & 72.7 \\
			Skip fine-tune on English & 71.0 & 73.6 & 72.3 \\
			Fine-tune on En/De mixed (CE) & \textbf{73.1} & 75.9 & 74.5 \\
			Fine-tune on En/De mixed (RW) & 65.5 & 42.6 & 51.6 \\
			\toprule
		\end{tabular}
	\end{center}
	\caption{Ablation study results evaluated on CoNLL2002 German NER data. All experiments used the mBERT-base model. RW denotes the re-weighted loss we proposed in the paper; CE denotes the regular cross entropy loss.}
	\label{tab:ablation}
\end{table}

From Table~\ref{tab:ablation}, we see that both pretraining on English and fine-tuning with pseudo German data are essential to get the best score. The RW loss performed better in sequential fine-tuning than in simultaneous training with mixed English and German data, this is probably because noise portion in English training set is much smaller than in the German pseudo-labeled training set, using RW loss on English data failed to exploit the fine-grained information in some hard examples and results in insufficiently optimized model. Another observation is that training the mBERT with a combination of English and German using cross entropy loss, we can get almost the same score as our best model, which is trained with two stages.

\section{Conclusion}
In this paper, we proposed a new method of doing NER cross-lingual transfer with parallel corpus. By leveraging the XLM-R model for entity projection, we are able to make the whole pipeline automatic and free from human engineered feature or data, so that it could be applied to any other language that has a rich resource of translation data without extra cost. This method also has the potential to be extended to other NLP tasks, such as question answering. In this paper, we thoroughly tested the new method in four languages, and it is most effective in Chinese. We also discussed the impact of parallel data domain on NER transfer performance and found that a combination of different domain parallel corpora yielded the best average results. We also verified the contribution of the pseudo-labeled parallel data by an ablation study. In the future we will further improve the alignment model precision and also explore alternative ways of transfer such as self teaching instead of direct fine-tuning. We are also interested to see how the propose approach generalizes to cross-lingual transfers for other NLP tasks.

\bibliography{ner_acl}
\bibliographystyle{acl_natbib}

\end{document}